\def\year{2020}\relax
\documentclass[letterpaper]{article} % DO NOT CHANGE THIS
\usepackage{aaai20}  % DO NOT CHANGE THIS
\usepackage{times}  % DO NOT CHANGE THIS
\usepackage{helvet} % DO NOT CHANGE THIS
\usepackage{courier}  % DO NOT CHANGE THIS
\usepackage[hyphens]{url}  % DO NOT CHANGE THIS
\usepackage{graphicx} % DO NOT CHANGE THIS
\usepackage{graphicx}  % DO NOT CHANGE THIS
\usepackage{booktabs}       % professional-quality tables
\usepackage{algorithm}
\usepackage{algorithmic}
\usepackage{graphicx}
\usepackage{amsmath}
\usepackage{amssymb}

\title{A Constructive Approach for Data-Driven Randomized Learning of Feedforward Neural Networks}
\author{Grzegorz Dudek	
	\\  % All authors must be in the same font size and format. Use \Large and \textbf to achieve this result when breaking a line
Department of Electrical Engineering, Czestochowa University of Technology\\
42-200 Czestochowa, Poland\\
dudek@el.pcz.czest.pl % email address must be in roman text type, not monospace or sans serif
}
 
\begin{document}

\maketitle

\begin{abstract}
Feedforward neural networks with random hidden nodes suffer from a problem with the generation of random weights and biases as these are difficult to set optimally to obtain a good projection space. Typically, random parameters are drawn from an interval which is fixed before or adapted during the learning process. Due to the different functions of the weights and biases, selecting them both from the same interval is not a good idea. Recently more sophisticated methods of random parameters generation have been developed, such as the data-driven method proposed in \cite{Anon19}, where the sigmoids are placed in randomly selected regions of the input space and then their slopes are adjusted to the local fluctuations of the target function. In this work, we propose an extended version of this method, which constructs iteratively the network architecture. This method successively generates new hidden nodes and accepts them if the training error decreases significantly. The threshold of acceptance is adapted to the current training stage. At the beginning of the training process only those nodes which lead to the largest error reduction are accepted. Then, the threshold is reduced by half to accept those nodes which model the target function details more accurately. This leads to faster convergence and more compact network architecture, as it includes only "significant" neurons. Several application examples are given which confirm this thesis. 
\end{abstract}

\section{Introduction}
A feedforward neural network (FNN) with a single hidden layer containing a finite number of neurons is able to approximate any continuous functions under mild assumptions on the activation function (AF). Unfortunately, a learning process which uses gradient based algorithms is sensitive to the local minima of the error function as well as its flat regions and saddle points. Moreover, the gradient calculations are time consuming, especially for large network architectures, complex target functions (TFs) and big training data sets. In randomized learning, the optimization problem, which is highly non-convex when gradient descent algorithms are used, becomes convex \cite{Pri15}. This is because the parameters of the hidden nodes are not learned but are selected randomly and stay fixed. The only adaptation occurs in the output weights. As the optimization problem is brought to a linear form, the output weights can be learned using a standard linear least-squares. The learning process in such a case is easier to implement and much faster than when using gradient descent algorithms for the full learning of all parameters, i.e. weights and biases of the hidden and output nodes. Many simulation studies reported in the literature show high performance of the randomized neural models which is compared to fully adaptable ones.

The main problem in randomized learning is finding a way of generating the random hidden node weights and biases to obtain a good projection space. According to the standard approach the random weights and biases are selected randomly with a fixed interval from any continuous sampling distribution. It was proven that when the interval is symmetric, the FNN has a universal approximation property if the function to be approximated meets Lipschitz condition \cite{Hau95}.
The selection problem of the appropriate interval for parameters has not been solved as yet and is considered to be one of the major research challenges in the area of FNN randomized learning \cite{Zha16c}, \cite{Cao18}. In many practical applications the interval for random parameters is selected as $[-1,1]$ without any justification, regardless of the problem type (classification or regression), data, and AF type. Some works have shown that such an interval is misleading because the FNN is unable to model nonlinear maps, no matter how many training samples are provided or how many hidden nodes are used \cite{Li17}. So, to improve the performance in a specified task, the optimization of this interval is recommended \cite{Pao94}, \cite{Hau95}, \cite{Wan17}.

Noting that weights and biases have different functions (weights express slopes of the sigmoids and biases represent sigmoid positions or shifts), in \cite{Dud19} a method of selecting the random parameters was proposed which generates them separately, not both from the same interval. This method takes into account the data scope and type of the AFs. First, it adjusts the AF slopes to the TF and then shifts the AFs into the input space. Another way of generating random parameters was recently proposed in \cite{Dud19a}. In this approach the slope angles of the sigmoids are randomly generated from the interval adjusted to TF. Then the sigmoids are randomly rotated and shifted into the input hypercube. Both these methods performed very well on the TF with strong fluctuations, outperforming the standard approach with the fixed interval.

Another idea for generating random parameters, the data-driven method of randomized FNN learning (D-DM), was recently proposed in \cite{Anon19}. According to this method, the hidden node weights are not selected from a specified interval but, instead, the sigmoids building the fitted function (FF) are adjusted individually to the local complexity of the TF. To do so, first, the proposed method selects the input space regions (by randomly choosing the training points), and then places the sigmoids in these regions and adjusts the sigmoid slopes to the TF slopes in the neighborhoods of the chosen points. The linear combination of the sigmoids reflects the TF features in different regions of the input space. Simulation studies have shown that this approach brings very good results in the
approximation of the complex TFs when compared to both the standard
fixed interval methods and new state-of-the-art methods proposed recently in the literature.

In this work, we propose a new constructive approach for the data-driven randomized learning of FNNs. This extended version of the D-DM constructs the network architecture iteratively by adding new hidden nodes. The candidate nodes are successively generated and they are accepted or rejected depending on the approximation error. If the error is reduced significantly over the assumed threshold, the node is accepted. The threshold of acceptance is adapted to the current training stage, accepting in the initial iterations only those nodes which approximate the TF roughly. In the next iterations the threshold is successively reduced by half to accept nodes which lead to a more accurate modeling of the TF details. The final architecture of the resulting network is compact as it includes only "significant" nodes, which are necessary to construct the well-fitted function to the TF. Compared to other recently proposed constructive algorithms for randomized FNN learning \cite{Wan17}, \cite{Dai19}, our proposed method does not search for the optimal interval for the random parameters. Instead, the weights and biases of nodes are determined by the TF slopes in the randomly selected input regions. So, the resulting FNN is more dependent on data and includes only nodes which are necessary to model the TF details with a required accuracy.   

\section{Constructive Approach for Data-Driven Randomized Learning}
A FNN with a single hidden layer, including sigmoids as AFs, is considered. D-DM adjusts the sigmoids individually to the local complexity of the TF \cite{Anon19}. It places the sigmoids into randomly selected input regions and adjusts their slopes to the TF slopes in these regions. The algorithm firstly randomly selects the input space regions by drawing points $\mathbf{x}^*$ from the training set. Then, at each point $\mathbf{x}^*$ the sigmoid $S$ is located in such a way that it has one of its inflection points, $P$, in $\mathbf{x}^*$. The sigmoid value in the inflection point is $0.5$, thus:  

\begin{equation}
h(\mathbf{x}^*) = \frac{1}{1 + \exp\left(-\left(\mathbf{a}^T\mathbf{x}^* + b\right)\right)}=0.5
\label{eqSigM}
\end{equation}
where  $\mathbf{x}^*=[x_1^*, x_2^*,..., x_n^*]^T\subset \mathbb{R}^n$, $\mathbf{a}=[a_1, a_2,..., a_n]^T \subset \mathbb{R}^n$ is a vector of sigmoid weights and $b$ is a sigmoid bias.

The slope of sigmoid $S$ is adjusted to the TF slope in $\mathbf{x}^*$. To do so, the slope of the TF is approximated by hyperplane $T$, which is fitted to the neighborhood of $\mathbf{x}^*$. This neighborhood, $\Psi(\mathbf{x}^*)$, includes $\mathbf{x}^*$ and its $k$ nearest neighbors among the training points. Hyperplane $T$ is of the form:
\begin{equation}
y = a_1'x_1 + a_2'x_2+...+a_n'x_n +b'
\label{eqLin}
\end{equation}
where coefficient $a_j'$ expresses a slope of the hyperplane in $j$-th direction.  

If we assume that this hyperplane is tangent to sigmoid $S$ at $P$, then the partial derivatives of sigmoid $S$ and hyperplane $T$ must be the same in $P$, and we obtain the sigmoid weights (see \cite{Anon19} for details):

\begin{equation}
a_j = 4a_j', \quad  j = 1, 2, ..., n
\label{eqDer4a}
\end{equation}   

Having the weights $a_j$ we can calculate the bias of sigmoid $S$ based on the equation we obtain from \eqref{eqSigM}:

\begin{equation}
b = -\mathbf{a}^T\mathbf{x}^*
\label{eqDer5a}
\end{equation}

Note the completely different way of generating the hidden node parameters in D-DM when compared to the standard way. The weights $a_j$ are not randomly generated from a fixed interval but correspond to the TF local slope. The biases are also not randomly selected from the fixed interval, but are calculated as linear combinations of points $\mathbf{x}^*$ and sigmoid weights $\mathbf{a}$. They express the shifting of the sigmoids to the randomly selected input regions.     

After randomly selecting $m$ points $\mathbf{x}^*$ we generate a set of sigmoids reflecting the local features of the TF in different regions. These sigmoids are the basis functions which are linearly combined to obtain the FF. The least squares estimate of the weights in this combination are calculated as follows:

\begin{equation}
\boldsymbol{\beta} = \mathbf{H}^+\mathbf{Y}
\label{eq5}
\end{equation}
where $ \boldsymbol{\beta} = [\beta_1, \beta_2, \ldots, \beta_m]^T $ is a~vector of output weights, $ \mathbf{Y} = [y_1, y_2, \ldots, y_N]^T $ is a~vector of target outputs and $ \mathbf{H}^+ $ is the the Moore–Penrose pseudoinverse of matrix $ \mathbf{H} $, which is an $N$-by-$m$ hidden layer output matrix for all $N$ training samples and $m$ nodes.

As reported in \cite{Anon19}, D-DM produces very good results in the
approximation of the complex target functions when compared to the standard
method and new methods proposed recently in the literature. This is because it places the steepest sigmoid fragments inside the input space and adjusts the sigmoid slopes to the TF. So the saturated fragments of sigmoids are avoided in modeling the TF.

The constructive extension of D-DM (CD-DM) proposed in this work generates sigmoids in the randomly selected regions and adjust their slopes according to the algorithm described above. After generating a new node, an approximation error is calculated, and the node is accepted or rejected. It is accepted when the error reduction is over a threshold $\theta$. Otherwise it is rejected. The threshold $\theta$ adapts to the advancement of the construction process. Its initial value should be low enough to accept only the "key" nodes, which model roughly the shape of the TF. These nodes are searched randomly by generating candidate nodes, one by one, in different input space regions. The generation process stops when, during successive $Q$ iterations, no new node is accepted among the candidate nodes. Then, the threshold $\theta$ is halved, and new candidate nodes are generated. At the new value of the acceptance threshold, those nodes which model the TF in more detail are accepted. If after $Q$ iterations, no node is accepted, the threshold $\theta$ is halved again. This process of generating candidate nodes and reducing the threshold by half is continued until the assumed number of nodes $m$ is reached.

At each successive level of the threshold $\theta$ new neurons can be added to the network, which ensures a more and more accurate fitting of the FF to the TF. To avoid overfitting, the process should be stopped at the right moment. This moment is controlled by the total number of hidden nodes $m$, which should be selected in cross-validation.

In CD-DM the number of samples and their size do not affect the number of nodes. Only the TF complexity
affects the network size. TFs without fluctuations can be modeled using a small number of nodes. In such a case, new candidate nodes considered in the network construction process do not reduct the error over threshold $\theta$ and are rejected. In the case of the complex TF with fluctuations, new nodes are added in the fluctuation regions until they cause error reduction. If all TF fluctuations are modeled by a set of nodes, new candidate nodes are rejected. The level of fitting accuracy and bias–variance tradeoff are dependent on the final number of nodes $m$ and size of the neighborhood $k'$. These regularization parameters are adjusted in the cross-validation to prevent underfitting or overfitting.

The proposed CD-DM can be formulated in Algorithm 1.

\begin{algorithm}
	\caption{Constructive Approach for Data-Driven Randomized Learning of FNNs}
	\label{alg1}
	\begin{algorithmic}
		%\small
		\vspace{1mm}
		\STATE {\bfseries Input:}\\ 
		\vspace{1mm}
		\hspace{4mm} Number of hidden nodes $m$\\
		\hspace{4mm} Number of nearest neighbors $k\geq n$\\
		\hspace{4mm} Initial value of the threshold for the error change $\theta \leq 0$\\
		\hspace{4mm} Threshold for the number of successive iterations\\
		\hspace{8mm} without a node accepted $Q$\\
		\hspace{4mm} Training set \\
		\hspace{8mm} $ \Phi = \left\lbrace (\mathbf{x}_l, y_l) | \mathbf{x}_l \in \mathbb{R}^n, y_l \in \mathbb{R}, l = 1, 2,\ldots, N \right\rbrace $\\
		
		\vspace{1mm}
		\STATE {\bfseries Output:}\\ 
		\vspace{1mm}
		\hspace{4mm} Weights $ \mathbf{A} = \left[
		\begin{array}{ccc}
		a_{1,1} & \ldots & a_{m,1} \\
		\vdots & \ddots & \vdots \\
		a_{1,n} & \ldots & a_{m,n}
		\end{array}
		\right]	$  \\
		\hspace{4mm} Biases $ \mathbf{b} = [b_1, \ldots, b_m] $ \\    
		\vspace{1mm}
		\STATE {\bfseries Procedure:}\\
		\vspace{1mm}
		
		\STATE $i=1$, $q=1$, $RMSE_0=1$, $\mathbf{H}=[]$
		\WHILE{$i\leq m$}
		\STATE (a) Choose randomly $\mathbf{x}^* = \mathbf{x}_l \in \Phi $, \\ 
		\hspace{4mm} where $l \sim U\{1, 2, \ldots, N\}$ \\
		\STATE (b) Create the set $\Psi(\mathbf{x}^*)$ containing $\mathbf{x}^*$ and its $k$ nearest\\
		\hspace{4mm} neighbors in $\Phi $
		\STATE (c) Fit the hyperplane to $\Psi(\mathbf{x}^*)$:\\ 
		\hspace{4mm} $y = a_1'x_1 + a_2'x_2+...+a_n'x_n +b'$
		\STATE (d) Compute the weights for the $i$-th node:\\ 
		\hspace{4mm} $a_{i,j} = 4a_j', \quad  j = 1, 2, ..., n $
		\STATE (e) Compute the bias for the $i$-th node:\\
		\hspace{4mm} $b_i = -\sum\limits_{j=1}^n a_{i,j}x_j^*$
		\STATE (f) Add the column corresponding to the $i$-th node\\
		\hspace{4mm} to the hidden layer output matrix:
		
		\begin{equation*}
		\mathbf{H}' =
		\left[\mathbf{H}
		\begin{array}{c}
		1/\left(1 + \exp\left(-\left(\mathbf{a}_i^T\mathbf{x}_1 + b_i\right)\right)\right)\\
		\vdots \\
		1/\left(1 + \exp\left(-\left(\mathbf{a}_i^T\mathbf{x}_N + b_i\right)\right)\right)
		\end{array}
	\right]
	\end{equation*}
	
	\STATE (g) Compute the output weights for $i$ nodes:\\
	\hspace{4mm} $\boldsymbol{\beta} = \mathbf{H}'^+\mathbf{Y}$
	
	\STATE (h) Compute the network output for the training set $ \Phi $:\\
	\hspace{4mm} $\mathbf{Y}' = \mathbf{H}'\boldsymbol{\beta}$
	
	\STATE (i) Compute the network error:\\
	\hspace{4mm} $RMSE_i = \sqrt{\frac{1}{N}\sum_{l=1}^{N}(y'_l - y_l)^2}$
	
	\STATE (j) Compute the error change:\\
	\hspace{4mm} $\Delta RMSE = RMSE_i - RMSE_{i-1}$
	
	\STATE (k) Accept or reject the $i$-th node:\\
	\hspace{4mm} $q=q+1$\\
	\hspace{4mm} \textbf{if} $\Delta RMSE \leq \theta$ \textbf{then}\\
	\hspace{8mm} $i=i+1$, $q=1$, $\mathbf{H}=\mathbf{H}'$\\
	\hspace{4mm} \textbf{end if}\\ 
	
	\STATE (l) Adapt the threshold for the error change:\\
	\hspace{4mm} \textbf{if} $q \geq Q$ \textbf{then}\\
	\hspace{8mm} $\theta=\theta/2$\\
	\hspace{4mm} \textbf{end if}\\ 
	
	\ENDWHILE
\end{algorithmic}
\end{algorithm}

The proposed method has four parameters, defined as follows:
\begin{itemize}
\item
$m$ -- the final number of hidden nodes. The algorithm adds successive nodes to the hidden layer if they reduce the error over the threshold $\theta$ until the final number of nodes $m$ is reached. Intuitively, more complex TF needs more hidden nodes to model accurately the function fluctuations. However, too many nodes leads to overfitting.      
\item
$k'$ -- size of the neighborhood including $k'=k+1$ training points, i.e. $\mathbf{x}^*$ and its $k$ nearest neighbors. The neighborhood $\Psi(\mathbf{x}^*)$ expresses the local features of the TF around the selected point $\mathbf{x}^*$. The sigmoid slope is determined on this set $\Psi(\mathbf{x}^*)$ to model the selected region of the TF. The optimal value of $k'$ depends on the noise level observed in the data, the TF complexity and the data density. A low value of $k'$ at a high level of noise leads to overfitting. On the other hand, too large $k'$ causes underfitting. The size of the neighborhood controls the bias-variance tradeoff of the model, as well as the number of nodes $m$.
\item
$\theta$ -- the threshold for the error change. This is the threshold of acceptance of any new hidden node which is added during the construction process. When, after adding a node, the reduction in error is over this threshold the node is accepted. Otherwise, it is rejected. If no node has been added for the subsequent $Q$ iterations, the threshold $\theta$ is halved. At successive levels of $\theta$ those new nodes which model the TF in more and more detail are accepted, up until the final number of nodes $m$ is reached. The initial value of $\theta$ should be low enough. When it is too low, it does not accept any nodes and increases rapidly in the first iterations of the construction process (so due to self-adaptation a too low initial value of $\theta$ is not a problem). When it is too large, it accepts most nodes without selection. This brings the proposed methods to the original D-DM.  
\item
$Q$ -- the threshold for the number of successive iterations without a node accepted. The method searches for the nodes to place them in the input space and so reduce the error above the threshold $\theta$. If such a node is not found by $Q$ iterations, the threshold $\theta$ is halved and the searching process is repeated, i.e. new candidate nodes are generated. Reaching the total number of $m$ nodes in the hidden layer completes the construction process.
\end{itemize}

\section{Simulation study}
\label{others}

In this section, to illustrate the proposed method and compare the performances of D-DM in its original and constructive versions, we report some experimental results over several regression problems. To evaluate the network performance we use root mean square error (RMSE). In all cases the thresholds for the error change and for the number of iterations without a node accepted were assumed to be fixed as: $\theta=-0.01$, $Q=50$. They were set on the basis of the preliminary simulations. The initial value of $\theta$ is not that important provided it is large enough. When it is too low, it does not accept any nodes and increases rapidly in the first iterations of the construction process. The simulations were carried out in the MATLAB 2018a environment, running an Intel i7-6950X 3.00 GHz processor, 48 GB RAM memory. All the results reported in this work take averages over 100 independent trials.

The first problem is a single-variable function approximation, where the TF is in the form:

\begin{equation}
g(x) = 0.2e^{-\left(10x - 4\right)^2} + 0.5e^{-\left(80x - 40\right)^2} + 0.3e^{-\left(80x - 20\right)^2}
\label{TF1}
\end{equation} 

The training dataset has $ 1000 $ points $ (x_l, y_l) $, where $ x_l $ are uniformly randomly distributed on $ [0, 1] $ and $ y_l $ are calculated from \eqref{TF1}. A test set has $ 300 $ points distributed regularly on $ [0, 1] $.

\begin{figure}
\centering
\includegraphics[width=0.4\textwidth]{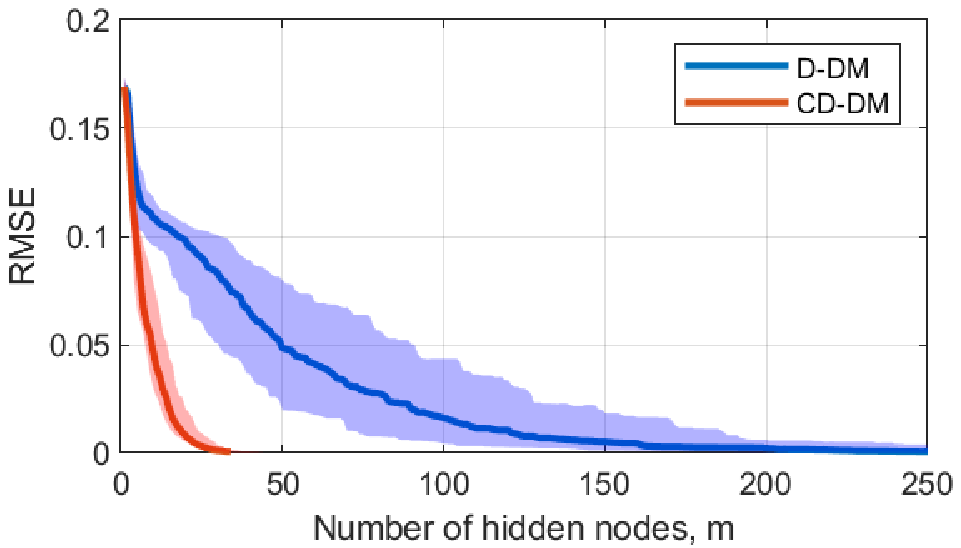}
\includegraphics[width=0.4\textwidth]{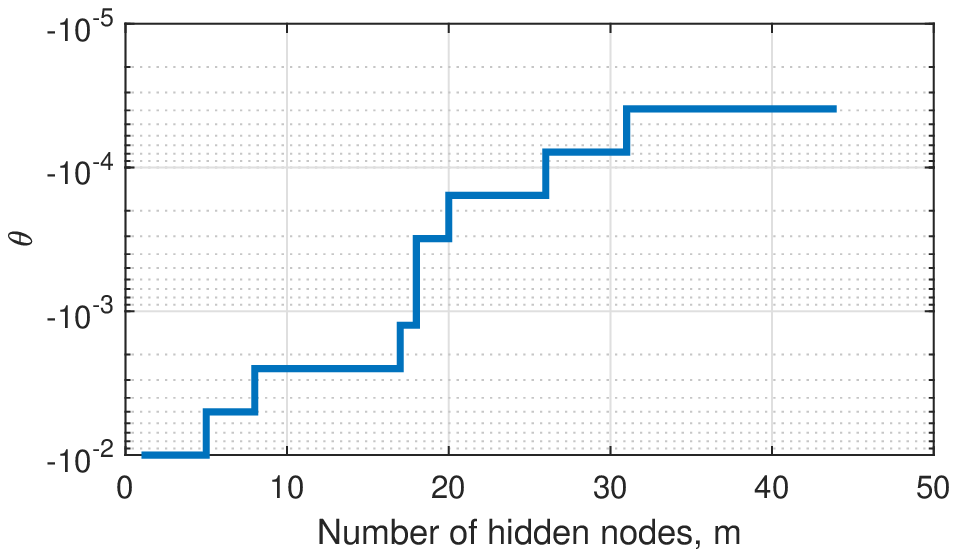}
\caption{TF \eqref{TF1} fitting: convergence (upper panel) and threshold $\theta$ (bottom panel).} 
\label{fig4}
\end{figure}

The upper panel of Fig. \ref{fig4} depicts the error on the test set for both versions of the data-driven method: original, D-DM, and constructive, CD-DM. The median curves over 100 runs are shown with the intervals between 10th and 90th percentiles. Note the much faster convergence of CD-DM. To get the same error level as D-DM with $250$ nodes ($RMSE\approx 0.00068$), CD-DM needs only $33$ nodes. Note also less dispersed results for CD-DM. The bottom panel of this figure shows the changes in the threshold $\theta$ during the construction process for one of the runs. As you can see from this figure, four nodes were added in the initial phase of the construction at $\theta = -0.01$, then $\theta$ was reduced by half and a further three nodes were added. After the next reduction in $\theta$ by half, nine nodes were added etc.

\begin{figure}
\centering
\includegraphics[width=0.4\textwidth]{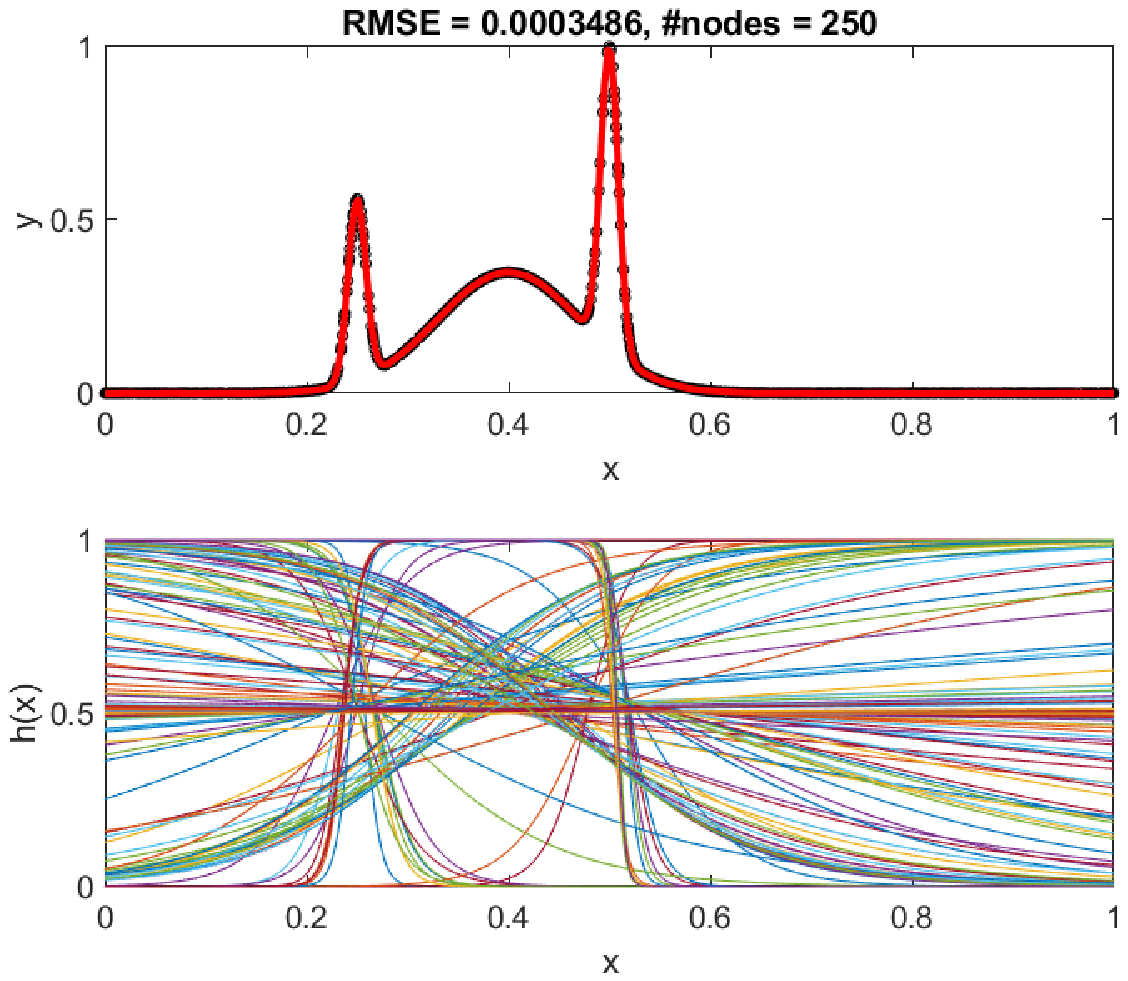}
\includegraphics[width=0.4\textwidth]{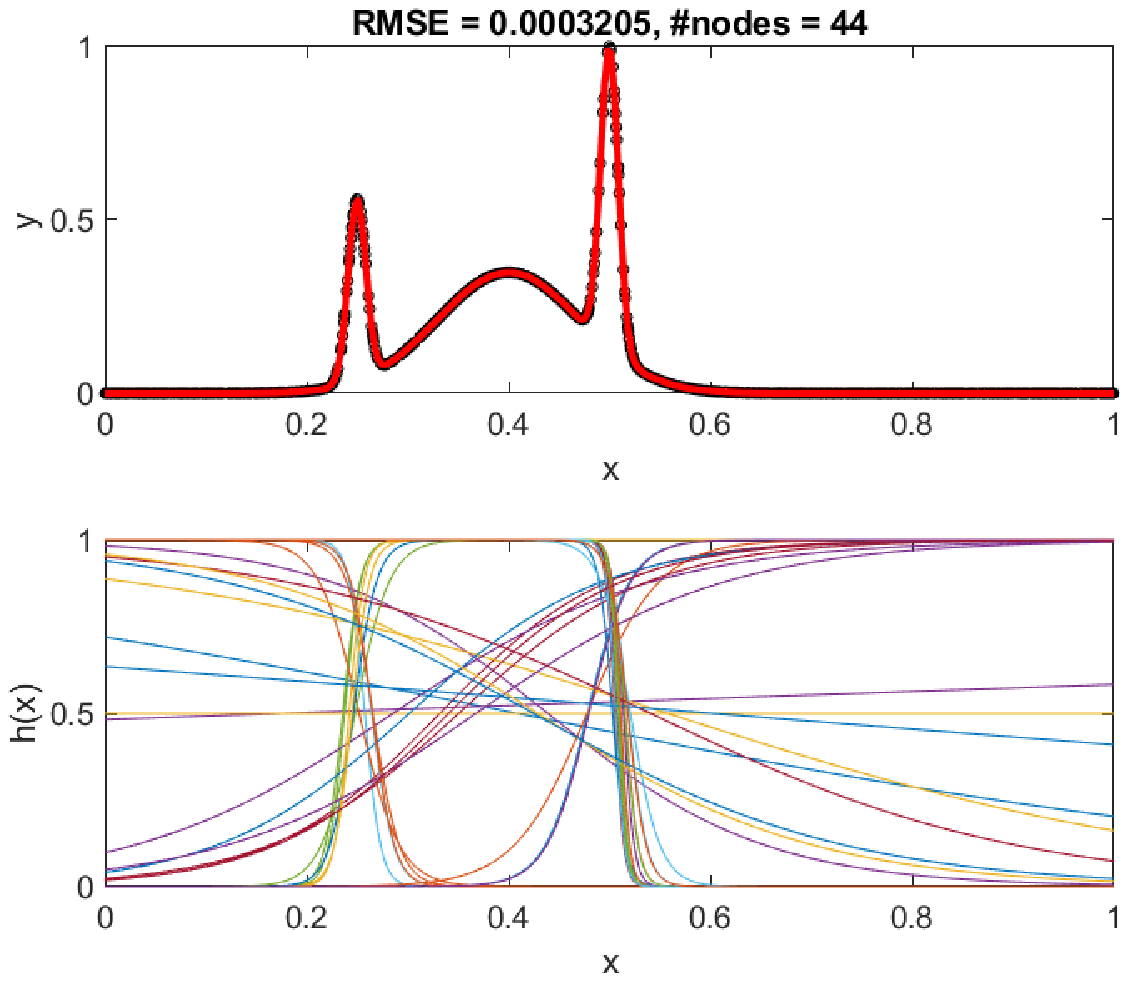}
\caption{TF \eqref{TF1} fitting: fitted curves and the sigmoids constructing them for D-DM (upper panel) and for CD-DM (bottom panel).} 
\label{fig5}
\end{figure}

The modeling performance is depicted in Fig. \ref{fig5} where the sigmoid $h(x)$ distributions in the input space are shown for both D-DM and CD-DM as well as the resulting fitted curves. Note the many flat nodes in the D-DM case. Introducing such flat neurons rarely decreases the error so CD-DM does not accept them during the FF construction process. Consequently, only $44$ nodes, usually those steep ones which correspond to the steep fragments of the function, are selected by CD-DM to approximate the TF. 

\begin{figure}
\centering
\includegraphics[width=0.49\textwidth]{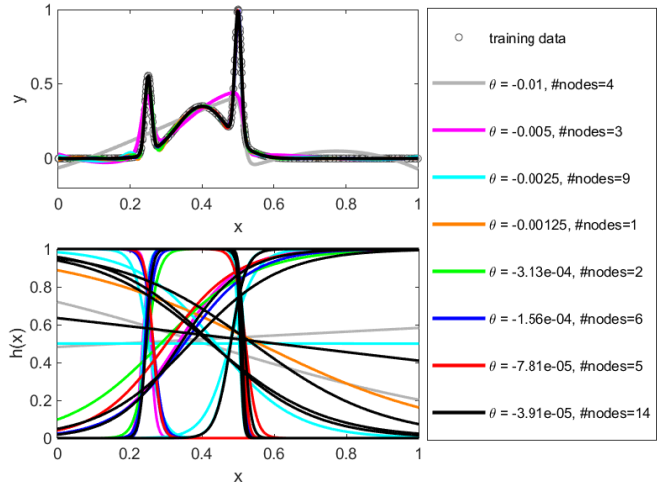}
\caption{TF \eqref{TF1} fitting: construction process of the fitted curve.} 
\label{fig5a}
\end{figure}

Fig. \ref{fig5a} demonstrates the construction process of the FF. The sigmoids selected at the successive $\theta$ levels are shown in different colors along with the fitted curves composed with them. The curve built from four nodes for the initial $\theta$ level is shown in gray. As you can see, it reflects only main features of the TF. Adding more nodes brings the curve closer to TF. At $\theta = -3.13\cdot 10^{-4}$ we have quite a good fit, except for those regions around the left spike where FF fluctuations are observed. Nodes added at the next $\theta$ levels smooth out these fluctuations.

The following examples of regression problems include a two variable function approximation and three real-world modeling tasks:

\begin{itemize}
\item 
Approximation of two variable TF:

\begin{equation}
g(\mathbf{x}) = \sin\left(20\cdot \exp\left(x_1\right)\right)\cdot x_1^2 + \sin\left(20\cdot \exp\left(x_2\right)\right)\cdot x_2^2
\label{TF2}
\end{equation}

The training set contains $ 5000 $ points $ (x_l, y_l) $, where components of $\mathbf{x}_l$ are independently uniformly randomly distributed on $ [0, 1] $ and $ y_l $ are generated from \eqref{TF2}, then normalized to the range $[0, 1]$ and distorted by adding the uniform noise distributed in $ [-0.2, 0.2] $. A test set of the same size is created in a similar way.

\item
Stock -- daily stock prices from January 1988 through October 1991, for ten aerospace companies. The task is to approximate the price of the 10th company given the prices of the others. There are $ 950 $ samples composed of nine input variables and one output variable.
\item 
Concrete -- the dataset contains the concrete's compressive strength, age, and seven ingredients. The task is to approximate the highly nonlinear relationship between the concrete's compressive strength and the ingredients and age. There are $ 1020 $ samples composed of eight input variables and one output variable.
\item 
Compactiv -- the Computer Activity dataset is a~collection of computer systems activity measures. The data was collected from a~Sun Sparcstation 20/712 with 128 Mbytes of memory running in a~multi-user university department. The task is to predict the portion of time that CPUs run in user mode. There are $ 8192 $ samples composed of $ 21 $ input variables (activity measures) and one output variable. 
\end{itemize}

The datasets Stock, Concrete and Compactiv were divided into training sets containing $ 75\% $ samples selected randomly, and test sets containing the remaining samples. The input and output variables are normalized into $ [0, 1] $.  These three datasets were downloaded from KEEL (Knowledge Extraction based on Evolutionary Learning) dataset repository (http://www.keel.es/). The neighborhood size $k'$ for D-DM was determined in 10-fold cross-validation. The same value of $k'$ was also assumed for CD-DM. The number of nodes was determined for these two method individually in cross-validation.     

\begin{figure}
\centering
\includegraphics[width=0.4\textwidth]{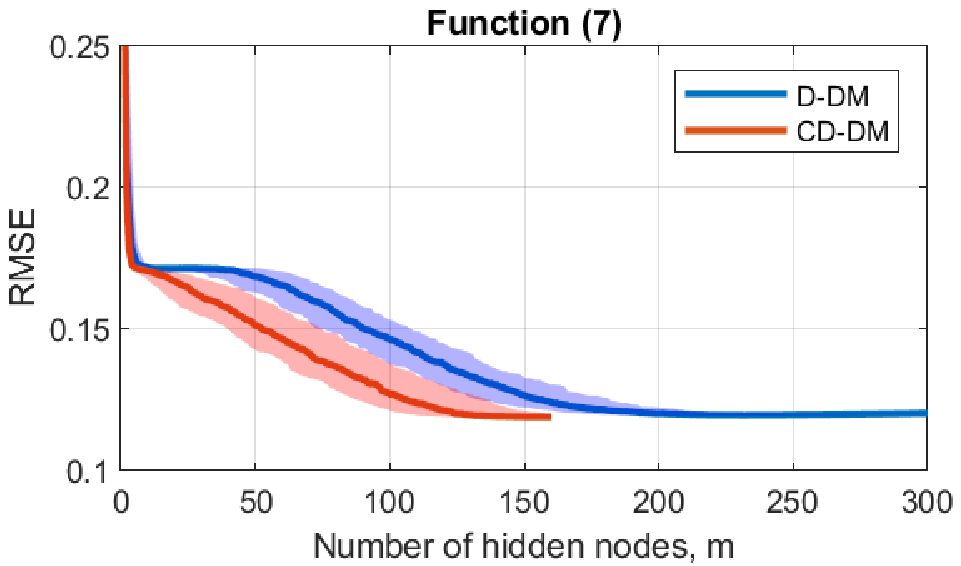}
\includegraphics[width=0.4\textwidth]{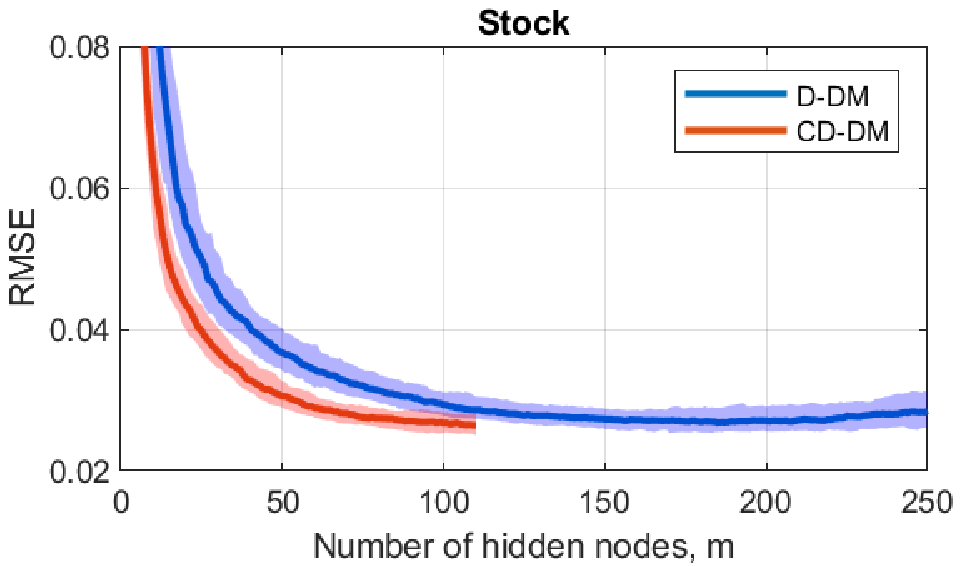}
\includegraphics[width=0.4\textwidth]{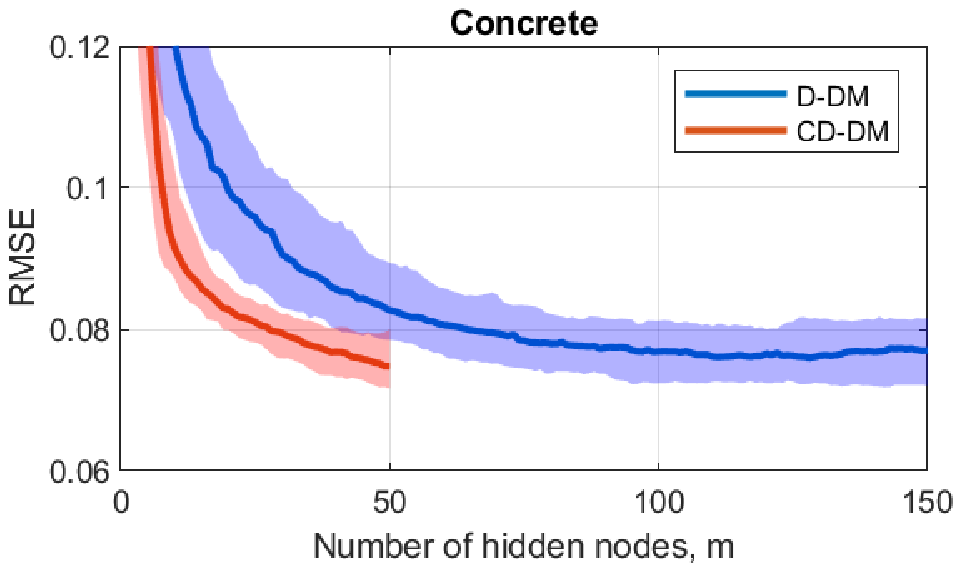}
\includegraphics[width=0.4\textwidth]{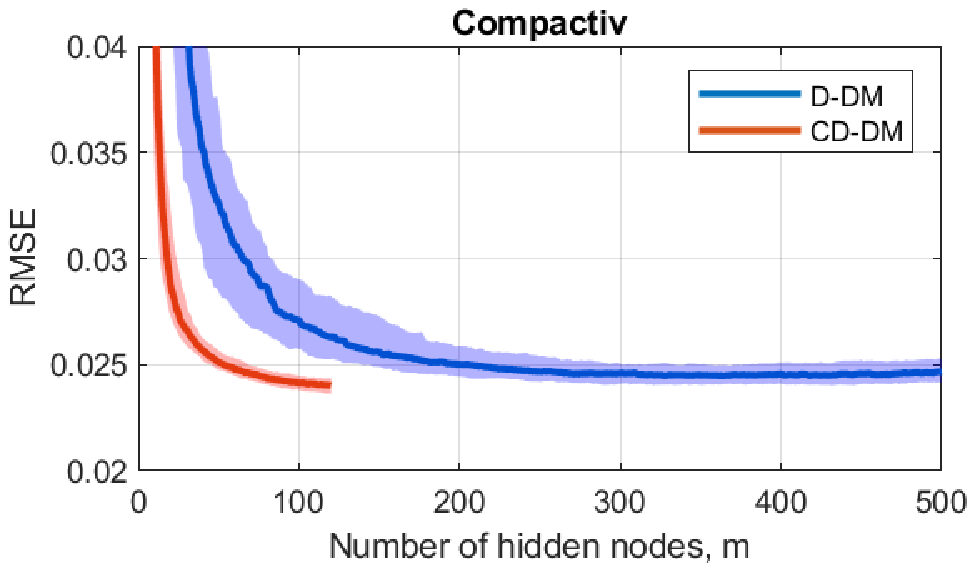}
\caption{Convergence of D-DM and the proposed CD-DM.} 
\label{fig6}
\end{figure}

\begin{figure}
\centering
\includegraphics[width=0.4\textwidth]{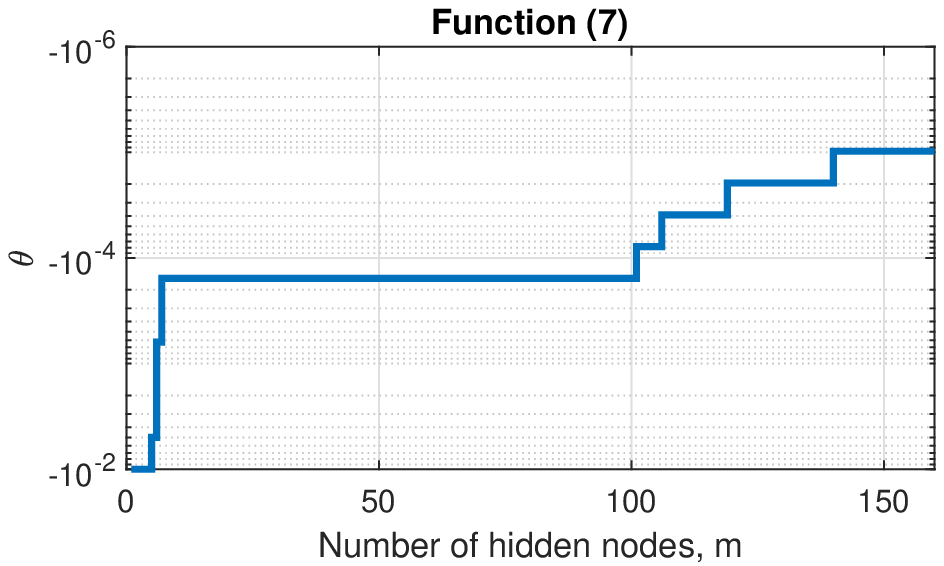}
\includegraphics[width=0.4\textwidth]{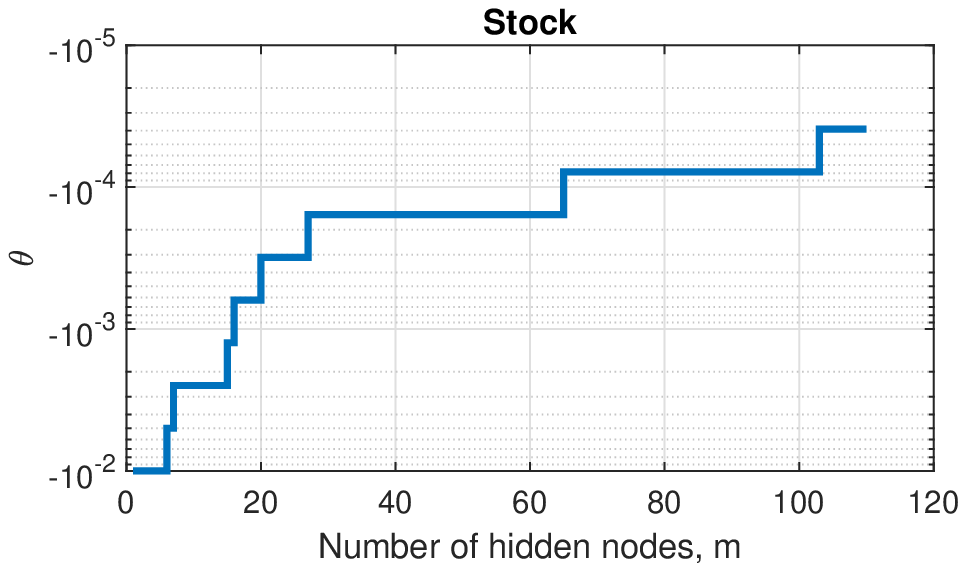}
\includegraphics[width=0.4\textwidth]{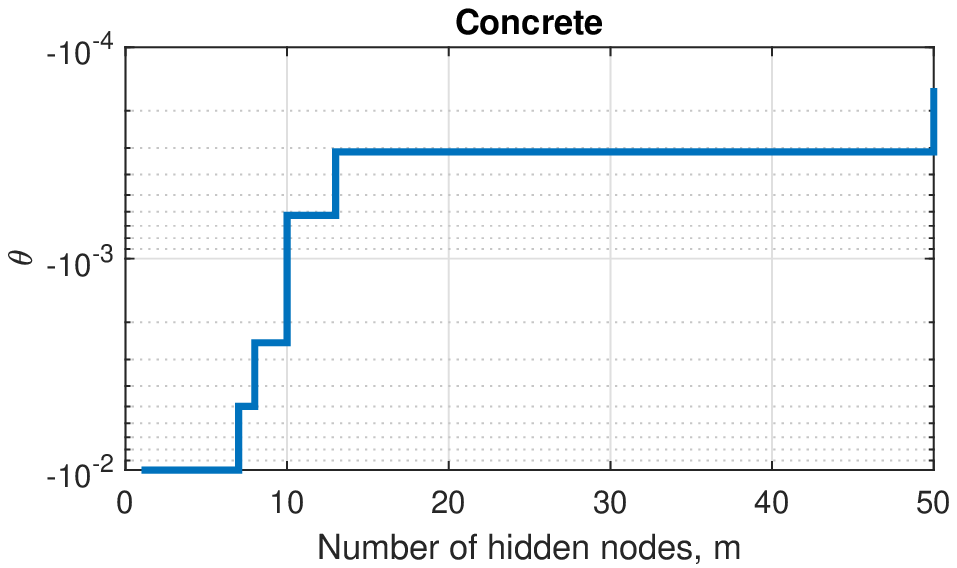}
\includegraphics[width=0.4\textwidth]{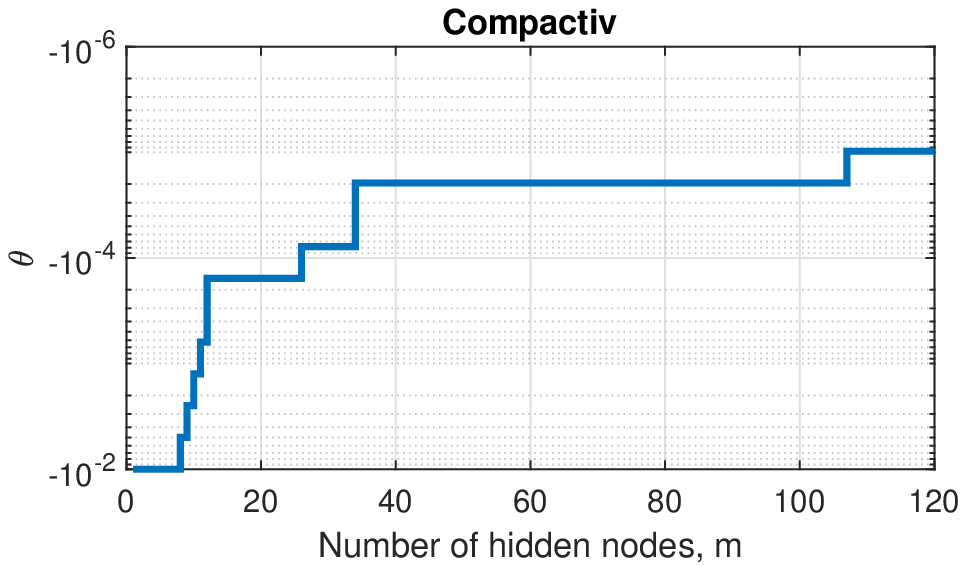}
\caption{Threshold $\theta$ in the proposed CD-DM.} 
\label{fig7}
\end{figure}

Fig. \ref{fig6} shows the convergence curves for D-DM and CD-DM. For each dataset, the proposed CD-DM converged much faster and achieved lower test RMSE than D-DM. Table 1 shows the errors and parameters of the models. For RMSE the medians and interquartile ranges over 100 runs are shown. Note that the number of hidden nodes for CD-DM is from two to five times smaller than for D-DM. The changes in threshold $\theta$ during the construction processes for one of the 100 runs are shown in Fig. \ref{fig7}. Note the rapid increase in the threshold in the first iterations and long segments without change in the threshold at its higher level, where many nodes which model the TF in more and more detail were added to the hidden layer.   

Taking into account the performance comparison reported in \cite{Anon19}, where the original D-DM are compared with the standard methods (with fixed and optimized intervals) as well as with two more sophisticated methods of random parameter generation recently proposed in the literature, \cite{Dud19}, \cite{Dud19a}, we can conclude that CD-DM outperforms standard and state-of-the-art methods in terms of accuracy, convergence speed and more compact network architecture.   

\begin{table*}[t]
\caption{Performance comparison of D-DM and CD-DM}
\label{sample-table}
\centering
\begin{tabular}{lllll}
	\toprule
	&\multicolumn{2}{c} {D-DM} 
	& \multicolumn{2}{c}{CD-DM}\\
	\cmidrule(r){2-3}
	\cmidrule(r){4-5}
	Data     & RMSE     & Parameters & RMSE & Parameters  \\
	\midrule
	Function \eqref{TF2} & $0.1204\pm 0.0003$  & $m=300, k'=35$ & $0.1191 \pm 0.0005$ & $m=160, k'=35, \theta = -0.01, Q = 50$     \\
	Stock & $0.0285 \pm 0.0028$ & $m=250, k'=30$ &$0.0265 \pm 0.0014$ &$m=110, k'=30, \theta = -0.01, Q = 50$ \\
	Concrete &$0.0770 \pm 0.0055$ & $m=150, k'=8$ &$0.0748 \pm 0.0034$ &$m=50, \ \ k'=8, \ \ \theta = -0.01, Q = 50$  \\
	Compactiv &$0.0247 \pm 0.0006$ & $m=500, k'=25$ &$0.0240 \pm 0.0003  $ &$m=120, k'=25, \theta = -0.01, Q = 50$ \\
	\bottomrule
\end{tabular}
\end{table*}

\section{Conclusion}

The key issue in FNNs with random hidden nodes is to generate the random weights and biases in such a way as to ensure good approximation properties of the network. The standard way of generating both parameters from the same fixed interval leads to weak performance, especially for complex target functions. The data-driven mechanism of randomized FNN learning proposed recently in \cite{Anon19} adjusts the random parameters to the local features of the target function and so improves the accuracy of fitting. First, the method randomly selects the input space regions by drawing the points from the training set. Then, the hyperplanes are fitted to the neighborhoods of the selected points and their coefficients are transformed into the sigmoid weights and biases. This results in the placement of the sigmoids in the selected regions of the input space and the adjustment of their slopes to local fluctuations in the target function. 

In this work, we propose a new constructive approach for data-driven FNN randomized learning. This extended version of the data-driven mechanism constructs iteratively the network architecture by adding successively new hidden nodes. These are accepted or rejected depending on the error. A node is accepted only if the error reduction is greater than the threshold. A low initial value of the threshold accepts only those nodes which ensure a rough function approximation. In the next steps the threshold is increased successively which leads to the acceptance of the nodes which more accurately model the details of the target function. To prevent overfitting the number of hidden nodes is limited to a value estimated in the cross-validation. As simulation research has shown, the proposed constructive method leads to faster convergence and more compact network architecture, as it includes only "significant" nodes.

Future work will focus on further analysis and improvement of the proposed method as well as other methods from this family, and their adaptation to classification problems.

\bibliographystyle{aaai}
\bibliography{references}

\end{document}